\documentclass{article}

\usepackage[dblblindworkshop, final]{neurips_2025}

\usepackage[utf8]{inputenc} 
\usepackage[T1]{fontenc}    
\usepackage{hyperref}       
\usepackage{url}            
\usepackage{booktabs}       
\usepackage{amsfonts}       
\usepackage{nicefrac}       
\usepackage{microtype}      
\usepackage{xcolor}         
\usepackage{amsmath}
\usepackage{subcaption}
\usepackage{graphicx}

\title{On Thin Ice: Towards Explainable Conservation Monitoring via Attribution and Perturbations}

\author{%
  Jiayi Zhou \\
  Duke University \\
  Durham, NC \\
  \And
  Günel Aghakishiyeva \\
  Duke University\\
  Durham, NC \\
  \And
  Saagar Arya \\
 Duke University \\
 Durham, NC \\
 \And
 Julian Dale \\
  Duke University\\
  Durham, NC \\
  \And
 James David Poling \\
 University of Agder  \\
 Grimstad, Norway \\
   \And
  Holly R. Houliston \\
 University of Cambridge  \\
  Cambridge, England \\
  \And
  Jamie N. Womble \\
  U.S. National Park Service  \\
  Department of Interior \\
  \And
  Gregory D. Larsen \\
  Alaska Spatial Science  \\
  Fairbanks, AK \\
\And
  David W. Johnston \\
 Duke University  \\
 Durham, NC \\
  \And
  Brinnae Bent \\
  Duke University \\
  Durham, NC \\
  \texttt{brinnae.bent@duke.edu} \\
}

\begin{document}

\maketitle

\begin{abstract}
  Computer vision can accelerate ecological research and conservation monitoring, yet adoption in ecology lags in part because of a lack of trust in black-box neural-network-based models. We seek to address this challenge by applying post-hoc explanations to provide evidence for predictions and document limitations that are important to field deployment. Using aerial imagery from Glacier Bay National Park, we train a Faster R-CNN to detect pinnipeds (harbor seals) and generate explanations via gradient-based class activation mapping (HiResCAM, LayerCAM), local interpretable model-agnostic explanations (LIME), and perturbation-based explanations. We assess explanations along three axes relevant to field use: (i) localization fidelity: whether high-attribution regions coincide with the animal rather than background context; (ii) faithfulness: whether deletion/insertion tests produce changes in detector confidence; and (iii) diagnostic utility: whether explanations reveal systematic failure modes. Explanations concentrate on seal torsos and contours rather than surrounding ice/rock, and removal of the seals reduces detection confidence, providing model-evidence for true positives. The analysis also uncovers recurrent error sources, including confusion between seals and black ice and rocks. We translate these findings into actionable next steps for model development, including more targeted data curation and augmentation. By pairing object detection with post-hoc explainability, we can move beyond “black-box” predictions toward auditable, decision-supporting tools for conservation monitoring.
\end{abstract}

\section{Introduction}

Computer vision is increasingly used to study wildlife populations \citep{gray2022drones}, habitat change \citep{xiong2024mangrove}, and the impacts of climate and human activity \citep{ruzicka2023semantic}, yet uptake in ecology remains uneven. While available technical expertise in ecological research teams is certainly a barrier to adoption, another critical one is trust: the most accurate detectors today are deep neural networks, whose internal reasoning is opaque to practitioners who must defend decisions about protected species and resource allocation \citep{gray2022drones,stavelin2021object,viegut2024detection,mpouziotas2024advanced}. A single false positive or negative can carry both operational and ethical costs, as overestimating a species’ presence may draw attention and resources away from truly at-risk populations, whereas underestimating it can hide early signs of decline and delay crucial conservation action \citep{chambert2015modeling,doull2021poacher}; accuracy alone is not sufficient. Ecological research and implementation require evidence for model decisions, visibility into failure modes, and a way to decide when to rely on automated predictions in the field \citep{gevaert2022explainable,buchelt2024exploring}.

Object detection in ecological imagery presents additional challenges for transparency. The targets are relatively small, partially occluded, and often visually confounded with background structures (e.g. dark ice or rock outcrops); animals appear in a variety of poses; and aerial acquisition often introduces scale variation and motion blur \citep{buchelt2024exploring}. In such settings, standard benchmarks can mask shortcut learning: a detector may key on context rather than the animal itself. Post-hoc explainability methods offer a path to audit these behaviors. Gradient-based class activation mapping can highlight influential regions in a model’s internal representations; model-agnostic perturbation methods can test whether predictions depend on putative evidence; and perturbation-based deletion/insertion tests can ask whether removing that evidence changes the decision. While bounding boxes from object detection models provide spatial localization, they do not reveal which features within the region drive the model’s prediction. Class activation maps, by contrast, can indicate whether the detector focuses on biologically meaningful structures rather than confounding background features. This distinction is crucial for fostering practitioner trust and for diagnosing edge cases where unusual pigmentation, occlusion, or environmental artifacts may lead the model to misinterpret the scene. However, most explainability work focuses on classification rather than detection, and, to our knowledge, there exist no studies that evaluate explanations against ecologically-relevant criteria in conservation workflows.

Our objective in this paper is to present evidence via a case study that post-hoc explainability should be routine when validating computer-vision models for ecology and conservation. Using a standard computer vision approach for pinniped (seal) monitoring, we apply CAM-style methods (HiResCAM, LayerCAM) \citep{draelos2021usehirescaminsteadgradcam,Jianglayercam}, LIME \citep{ribeiro2016whyitrustyoulime}, and perturbation-based explanations \citep{verma2022counterfactual}, and measure their value with field-relevant criteria: localization fidelity (do attributions align with the animal), faithfulness under deletion/insertion (do edits to the target object move detector confidence as expected), and diagnostic utility (do explanations reveal systematic errors). By moving beyond “black-box” predictions to auditable detections, our case study illustrates how explainability can bridge the trust gap for conservation monitoring. 

\section{Related Work}

Related work on explainable AI (XAI) is extensive within the broader AI community \citep{saarela2024recent}, yet crossover into ecology remains limited. Recent reviews in earth observation underscore emerging regulations and guidance that call for explainable algorithms and discuss their anticipated impact on practice \citep{gevaert2022explainable, buchelt2024exploring}. Despite this consensus, most ecological deep learning systems still operate as opaque “black boxes,” leaving gaps in understanding how models use visual evidence in ecological research and conservation workflows.

Early applications of XAI in ecology show promise but are concentrated in classification and regression tasks. For example, LIME has been used to provide localized insights in species distribution modeling \citep{ryo2021xai} and to highlight influential image regions for bird classification \citep{bird2025lime}. Spatially explicit explanations for object detection and segmentation are more established in other high-stakes domains: medical imaging uses them to reveal diagnostically relevant regions \citep{brima2024saliency}, and autonomous driving applies them to justify road detection and scene segmentation \citep{mankodiya2022odxai}. This disparity suggests an opportunity to adapt and evaluate such methods for ecological detection and monitoring tasks, where localization fidelity and faithfulness are critical.

The methodological foundation to bridge this gap is readily available through open-source toolkits \citep{jacobgil_gradcamlib,lime2016lib}. Building on this work, we apply spatially explicit explanations to ecological imagery and evaluate them against ecologically-relevant conservation criteria.

\section{Methods}
\subsection{General Pipeline}
We integrated an object detection model with post-hoc explainability methods to create a pipeline that interprets a ‘black-box’ model through clear visualizations and aids conservation monitoring decisions. Our dataset consists of 1,974 aerial drone images from Glacier Bay National Park, each annotated with bounding boxes for seals (Appendix A). We trained a detection model using Faster R-CNN, a widely applied model in ecological research \citep{Cipriano2025}, achieving a final mean average precision at 50\% intersection over union (mAP50) of 0.95 and mean average recall (mAR) of 0.69 on a held-out test set (Appendix B). We then analyzed the model using three explainability techniques (Appendix C): CAM-style methods (HiResCAM, LayerCAM) \citep{draelos2021usehirescaminsteadgradcam,Jianglayercam}, LIME \citep{ribeiro2016whyitrustyoulime}, and perturbation-based explanations \citep{verma2022counterfactual}. These explanations were evaluated against ecologically relevant criteria to assess localization fidelity, faithfulness, and diagnostic utility (Appendix D).

\subsection{Explainability Approaches}
Three explainability techniques are used to interpret the model's decision-making process, each offering a unique perspective on feature importance. First, Gradient-based CAM methods generate heatmaps that highlight image regions most influential to a specific prediction. We utilized two variants, HiResCAM \citep{draelos2021usehirescaminsteadgradcam} and LayerCAM \citep{Jianglayercam}, which offer higher spatial fidelity than traditional methods, a critical feature for localizing small or partially occluded animals in aerial imagery. These methods produce a spatial relevance map by analyzing the gradients and activations within the model's final convolutional block. Then, LIME explains predictions by approximating the model's local decision boundary using an interpretable surrogate model trained on perturbed input samples \citep{ribeiro2016whyitrustyoulime}. We adapted the standard LIME framework for the multi-instance object detection task with three improvements: improved superpixel segmentation, a detector-aware weighting scheme to aggregate explanations for multiple objects, and proximity-based spatial suppression to produce cleaner attribution maps. A perturbation-based approach is also used to assess which image features are necessary for detection. This method systematically and minimally alters an image to determine the smallest edit required to cause the model's prediction confidence to fall below a threshold and fail to detect an object. The process relies on a greedy forward-selection algorithm that iteratively perturbs the most important superpixels within or near the target's bounding box. Together, these three techniques demonstrate how to move from “black-box” predictions to explainable detections by providing visual insights into model behavior. Complete methods for the explainability approaches are provided in Appendix C. Code for explainability methods is available here: \href{https://github.com/duke-trust-lab/on-thin-ice}{https://github.com/duke-trust-lab/on-thin-ice}.

\subsection{Assessing Explanations Along Ecologically-Relevant Axes}
We evaluated explanations along three ecologically relevant dimensions: localization fidelity, faithfulness, and diagnostic utility. For localization fidelity, we looked at how well explanation methods focused on the animal rather than surrounding background. We measured the \textit{attribution ratio}, which is the proportion of high-attribution pixels within the annotated bounding boxes, and the \textit{maximum saliency hit-rate}, the percentage of images where the most important pixel fell inside a labeled box. Faithfulness considered whether explanations truly reflected the model’s reasoning, using our perturbation-based method to calculate two measures. The \textit{flip rate} measures the frequency of cases where removing the highlighted features caused the model to miss the target. The \textit{confidence drop} reflects the mean reduction in prediction confidence, both overall and for cases where the prediction was not successfully flipped. Diagnostic utility analyzed the practical value of explanations by reviewing false positive detections with saliency maps overlaid on ground-truth labels, allowing us to identify image features that led to false positive detections and suggest improvements to the model and dataset. Complete methods for explanation assessment are provided in Appendix D. Together, these measures reveal how well the explanations capture model behavior and guide trustworthy decision-making in conservation monitoring.

\section{Results}
We examine model predictions alongside post-hoc attributions to characterize correct predictions and uncover sources of error with the aim of providing explainable predictions that both increase trust in the model and inform model improvements. In a representative success (Figure 1, top row), both LayerCAM and HiResCAM reveal concentrated activation over the torsos and contours of the seals, while LIME highlights superpixels that overlap the same regions. This alignment supports localization fidelity. We quantify localization fidelity across 76 test images by quantifying the attribution ratio, or the normalized proportion of mid- to high-attribution pixels falling within annotated bounding boxes. LayerCAM achieved an attribution ratio of $67.69 \pm 25.13\%$, with a maximum-saliency hit rate of $94.70\%$ (125/132 boxes). HiResCAM achieved a comparable attribution ratio of $58.56 \pm 30.83\%$ and a hit rate of $69.70\%$ (92/132 boxes). LIME yielded an attribution ratio of $48.76 \pm 29.51\%$, with a maximum-saliency hit rate of $92.42\%$ (122/132 boxes). Visualizations, paired with quantitative metrics computed on the test dataset, provide an improved understanding as to whether detections are grounded in biologically meaningful regions or in the surrounding context. 

\begin{figure}[htbp]
  \centering
  \vspace{-0.5em}
  
  \begin{subfigure}[t]{0.21\textwidth}
      \includegraphics[width=\textwidth]{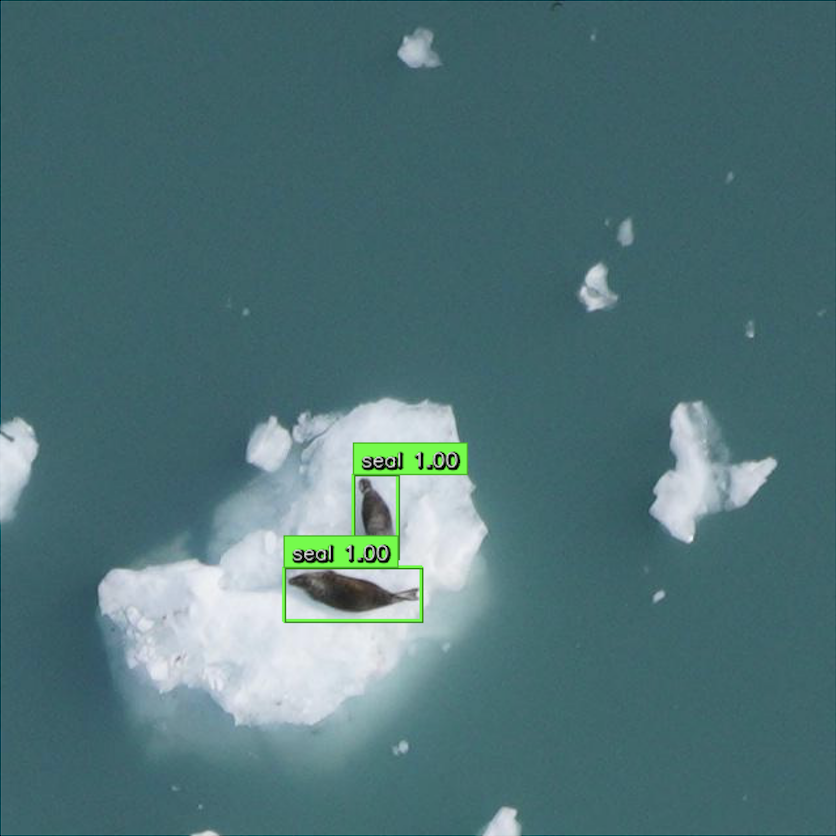}
      \caption{Original}
  \end{subfigure}
  \begin{subfigure}[t]{0.21\textwidth}
      \includegraphics[width=\textwidth]{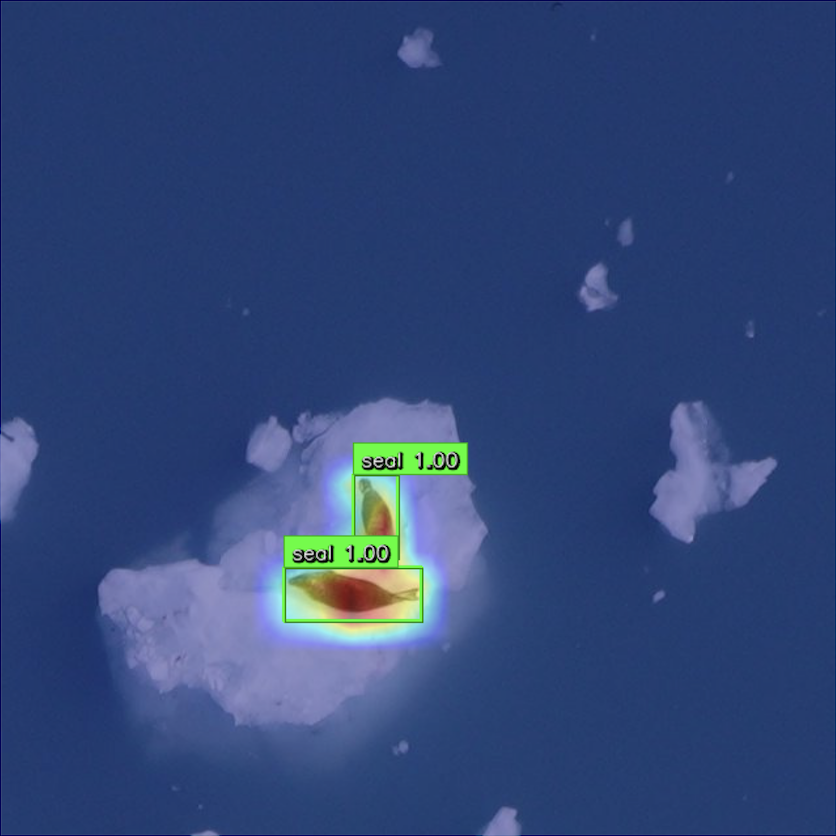}
      \caption{LayerCAM}
  \end{subfigure}
  \begin{subfigure}[t]{0.21\textwidth}
      \includegraphics[width=\textwidth]{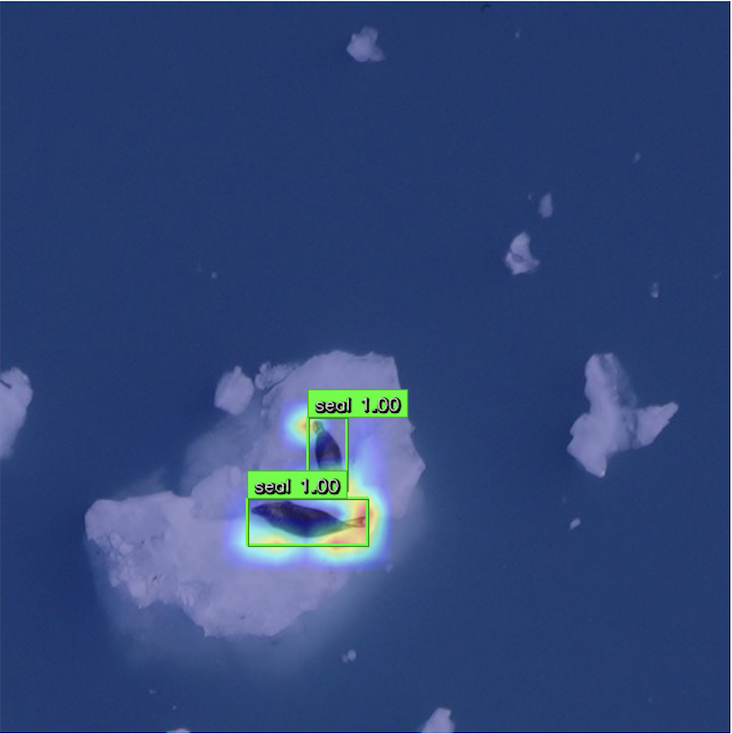}
      \caption{HiResCAM}
  \end{subfigure}
  \begin{subfigure}[t]{0.21\textwidth}
      \includegraphics[width=\textwidth]{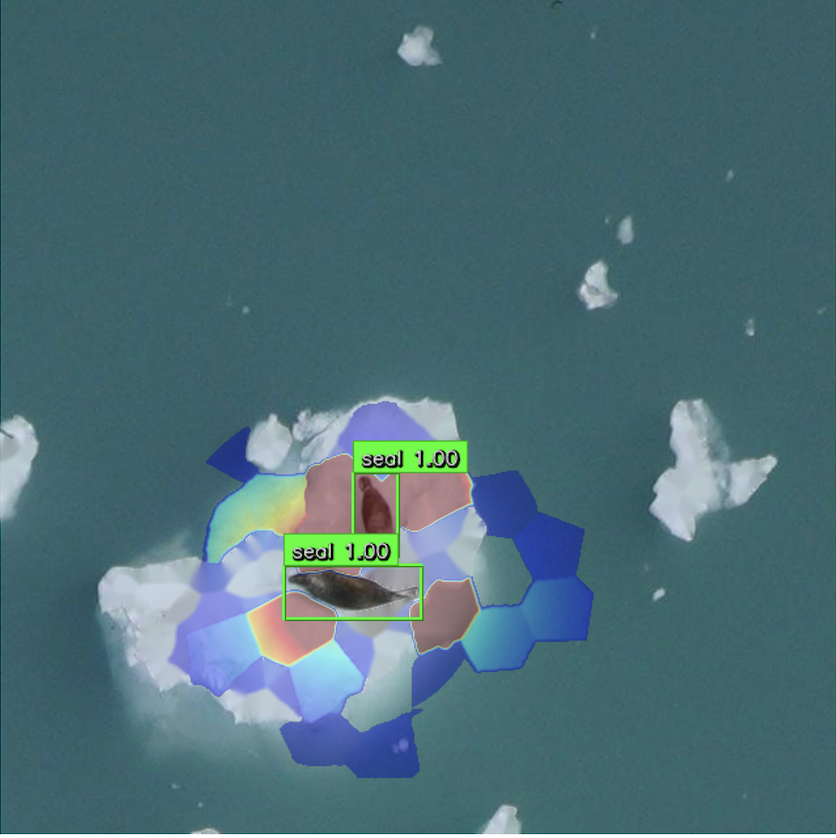}
      \caption{LIME}
  \end{subfigure}
  
  \begin{subfigure}[t]{0.21\textwidth}
      \includegraphics[width=\textwidth]{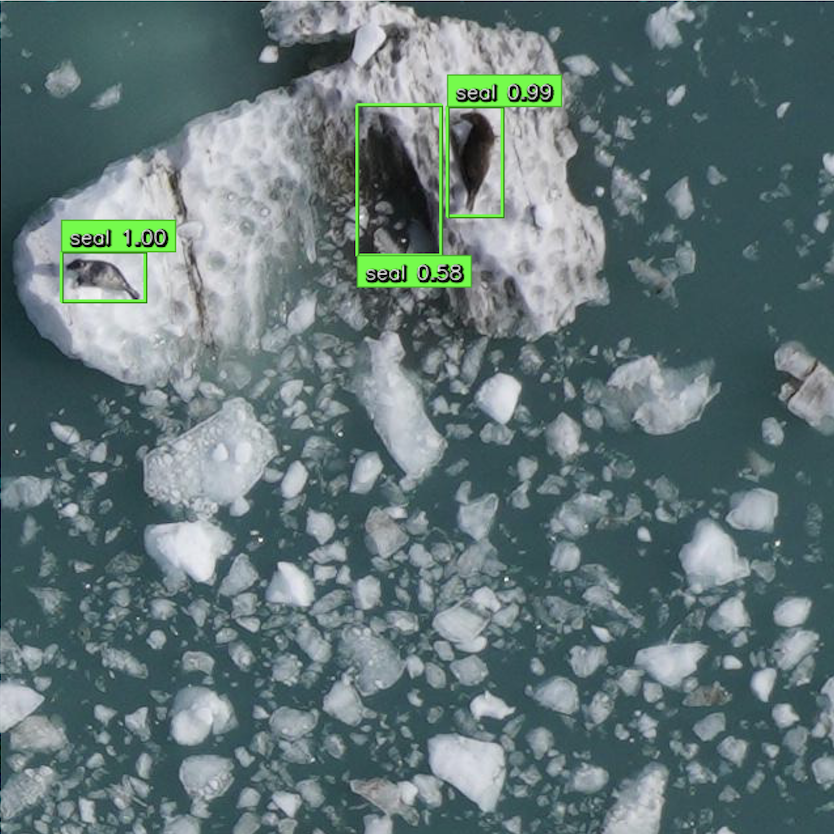}
      \caption{Original}
  \end{subfigure}
  \begin{subfigure}[t]{0.21\textwidth}
      \includegraphics[width=\textwidth]{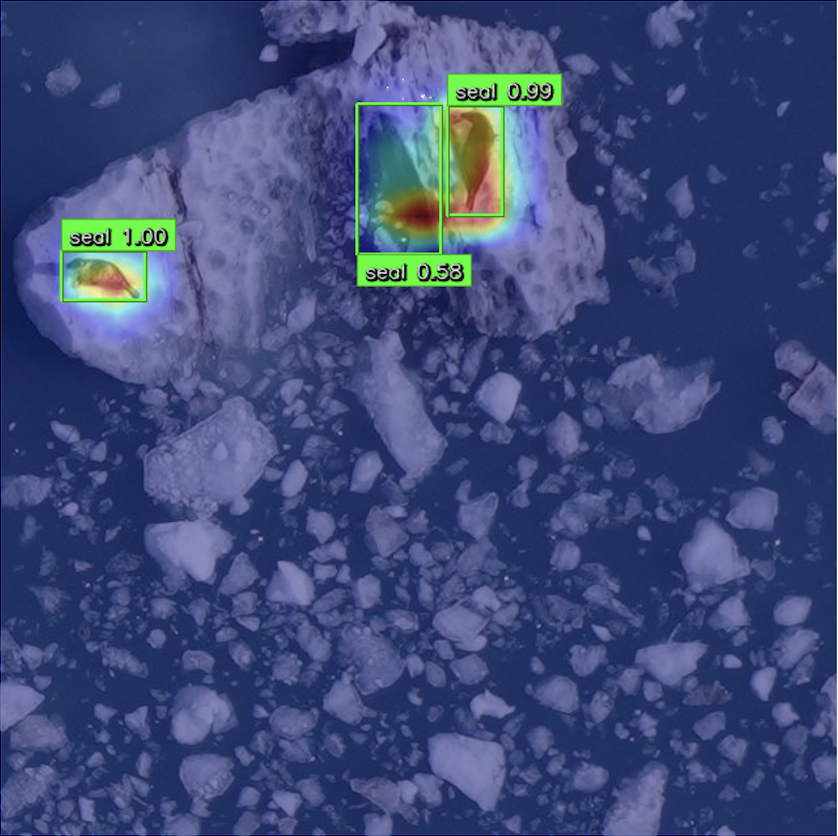}
      \caption{LayerCAM}
  \end{subfigure}
  \begin{subfigure}[t]{0.21\textwidth}
      \includegraphics[width=\textwidth]{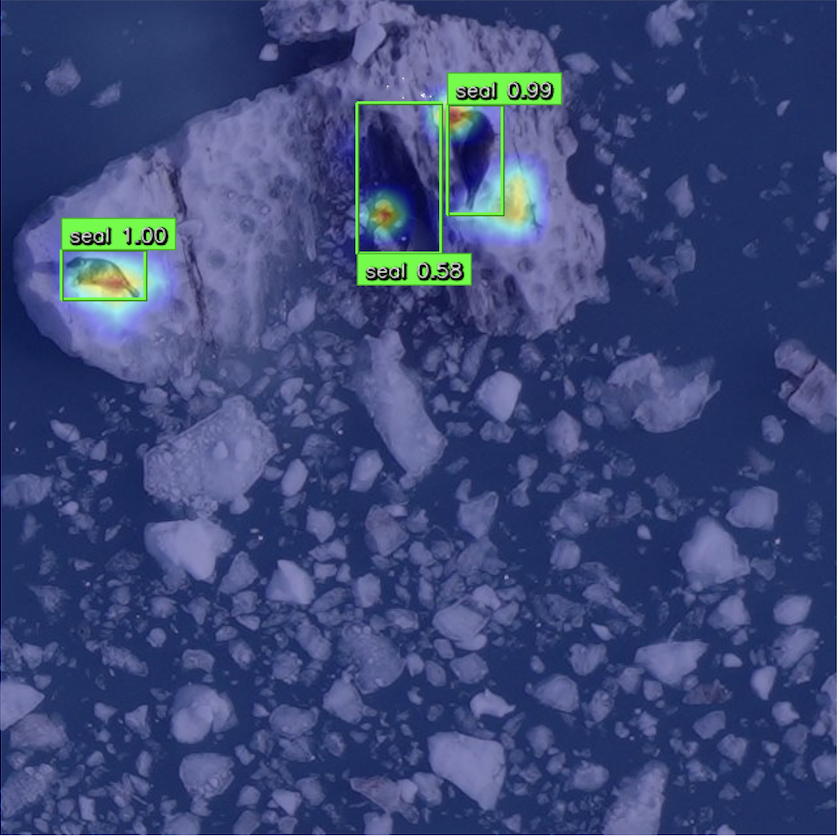}
      \caption{HiResCAM}
  \end{subfigure}
  \begin{subfigure}[t]{0.21\textwidth}
      \includegraphics[width=\textwidth]{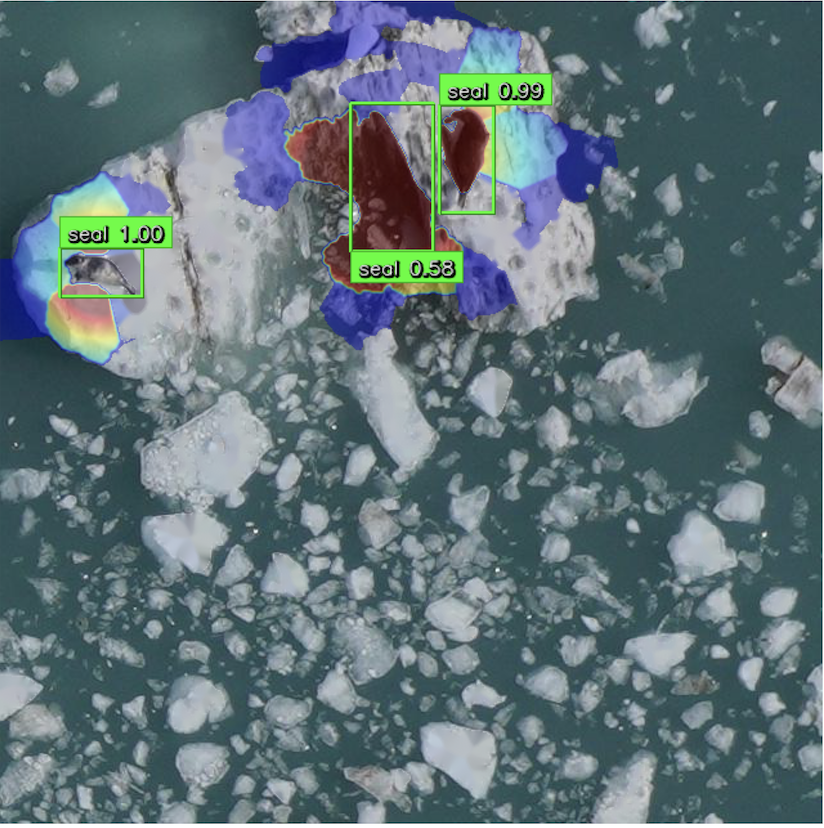}
      \caption{LIME}
  \end{subfigure}

  \caption{Visual explanations of harbor seal detections using Faster R-CNN. 
  Top row: Successful detection. In the attribution maps, a color spectrum represents relative importance, with warm colors (e.g., red, orange, yellow) indicating high contribution and cool colors (e.g., blue) indicating little or no contribution.
  (a) Original detections with bounding boxes. 
  (b) LayerCAM attribution (c) HiResCAM attribution (d) LIME attribution. 
  Bottom row: Challenging failure case where black ice is misidentified as a seal. 
  (e) Original detections with bounding boxes, including a false positive with 0.58 confidence. 
  (f) LayerCAM attribution (g) HiResCAM attribution (h) LIME attribution. The attribution maps reveal that the model confuses black ice with seals,
indicating a limitation in generalizing to environments with black ice or rocks.
  }
  
  \vspace{-0.5em}
\end{figure}

Attribution maps indicate where the detector attends, while perturbation-based explanations test whether those regions are necessary. In Figure 2, the seal is detected at full confidence; masking the seal body or replacing it with noise drives confidence to 0, whereas blurring reduces it to 0.77. Aggregated over 76 test images, mask and noise were most destructive, yielding mean confidence decreases of 0.97 and 0.87 with flip rates of 97.33\% and 80.00\%, respectively (flip = detection suppressed to zero). Blur caused a smaller average decrease (0.27) and a lower flip rate (21.33\%); when predictions did not flip, confidence fell by only 0.07 under blur (0.10 for mask; 0.37 for noise). Taken together, these results show that our model relies primarily on intact body contours and coherent feature patterns: occlusion or structure-breaking noise reliably removes detections, while blurred images retain sufficient morphology to sustain many true positives.

\begin{figure}[htbp]
  \centering
  \vspace{-0.5em}
  \begin{subfigure}[b]{0.33\textwidth}
        \includegraphics[width=\textwidth]{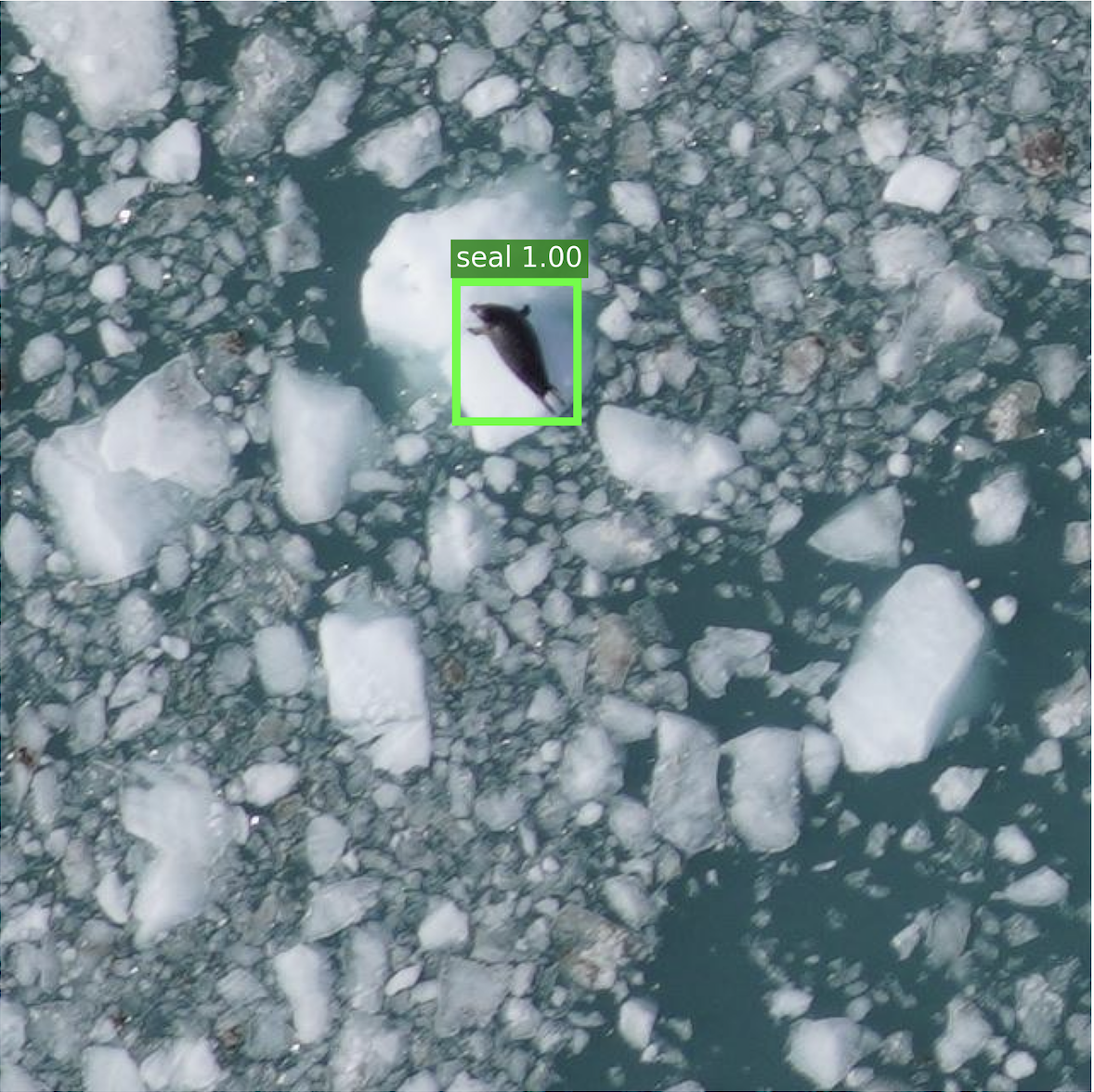}
        \caption{Original detection}
    \end{subfigure}
    \begin{subfigure}[b]{0.33\textwidth}
        \includegraphics[width=\textwidth]{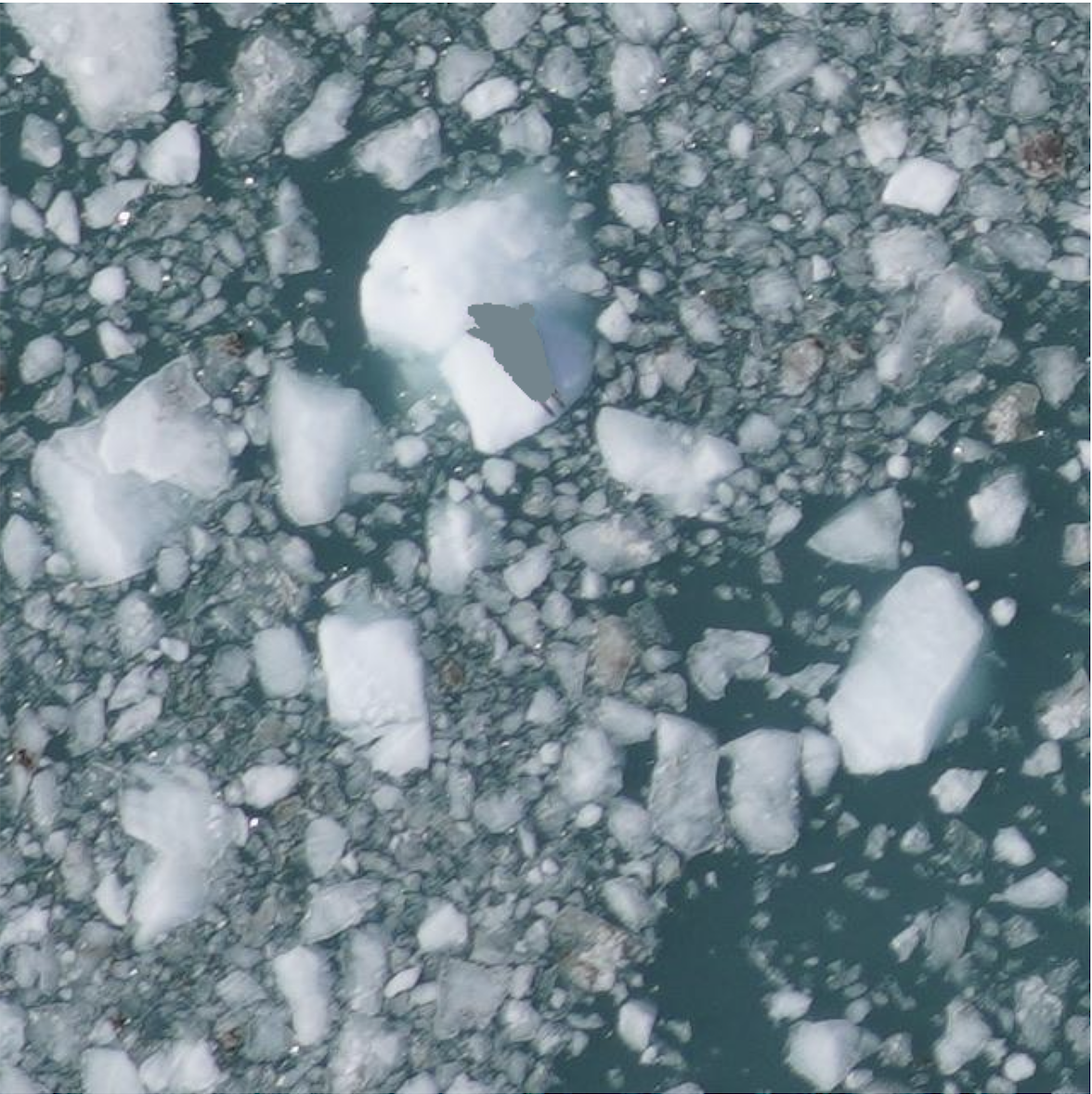}
        \caption{Mask}
    \end{subfigure}
    \\
    \begin{subfigure}[b]{0.33\textwidth}
        \includegraphics[width=\textwidth]{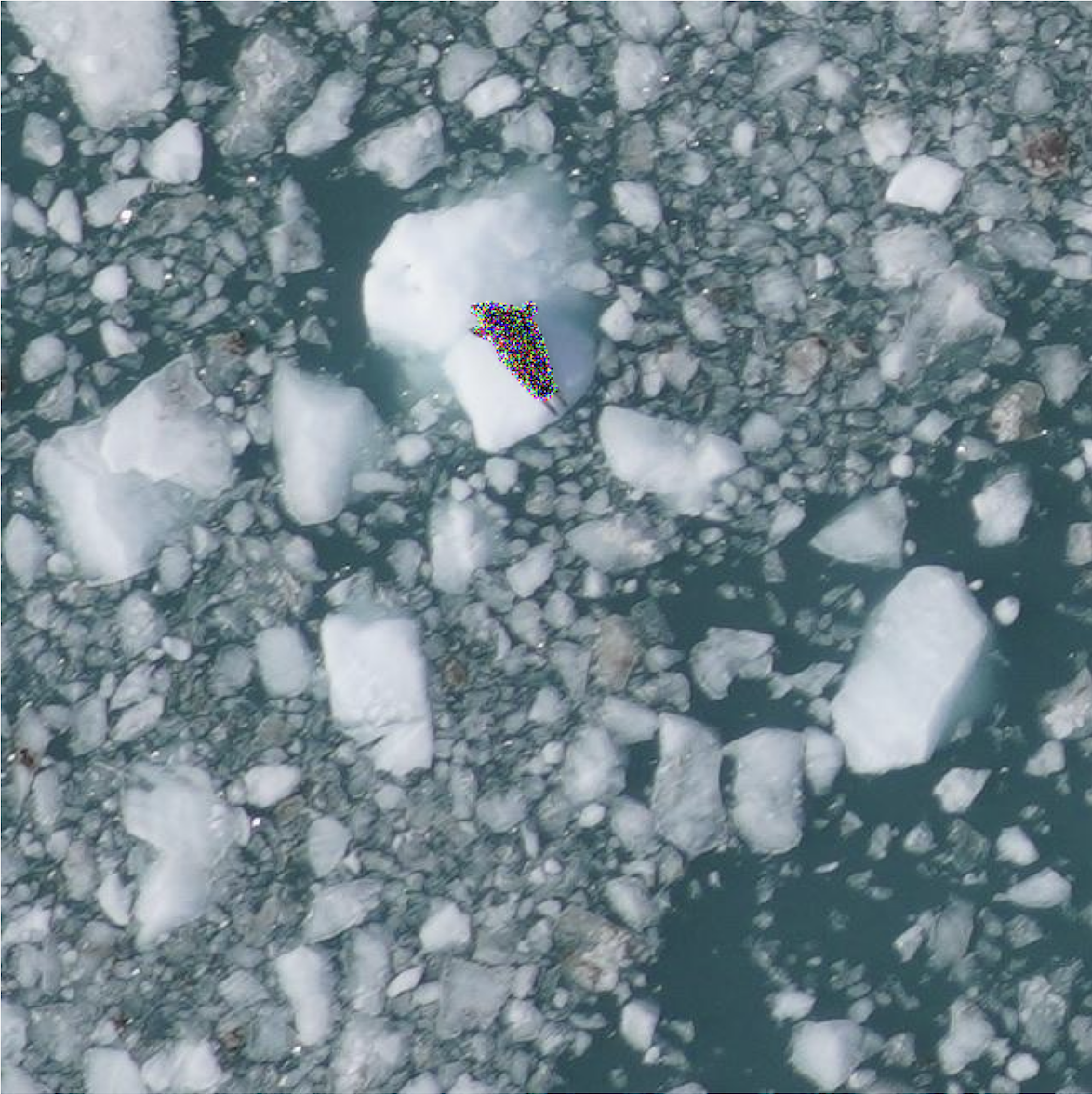}
        \caption{Noise}
    \end{subfigure}
    \begin{subfigure}[b]{0.33\textwidth}
        \includegraphics[width=\textwidth]{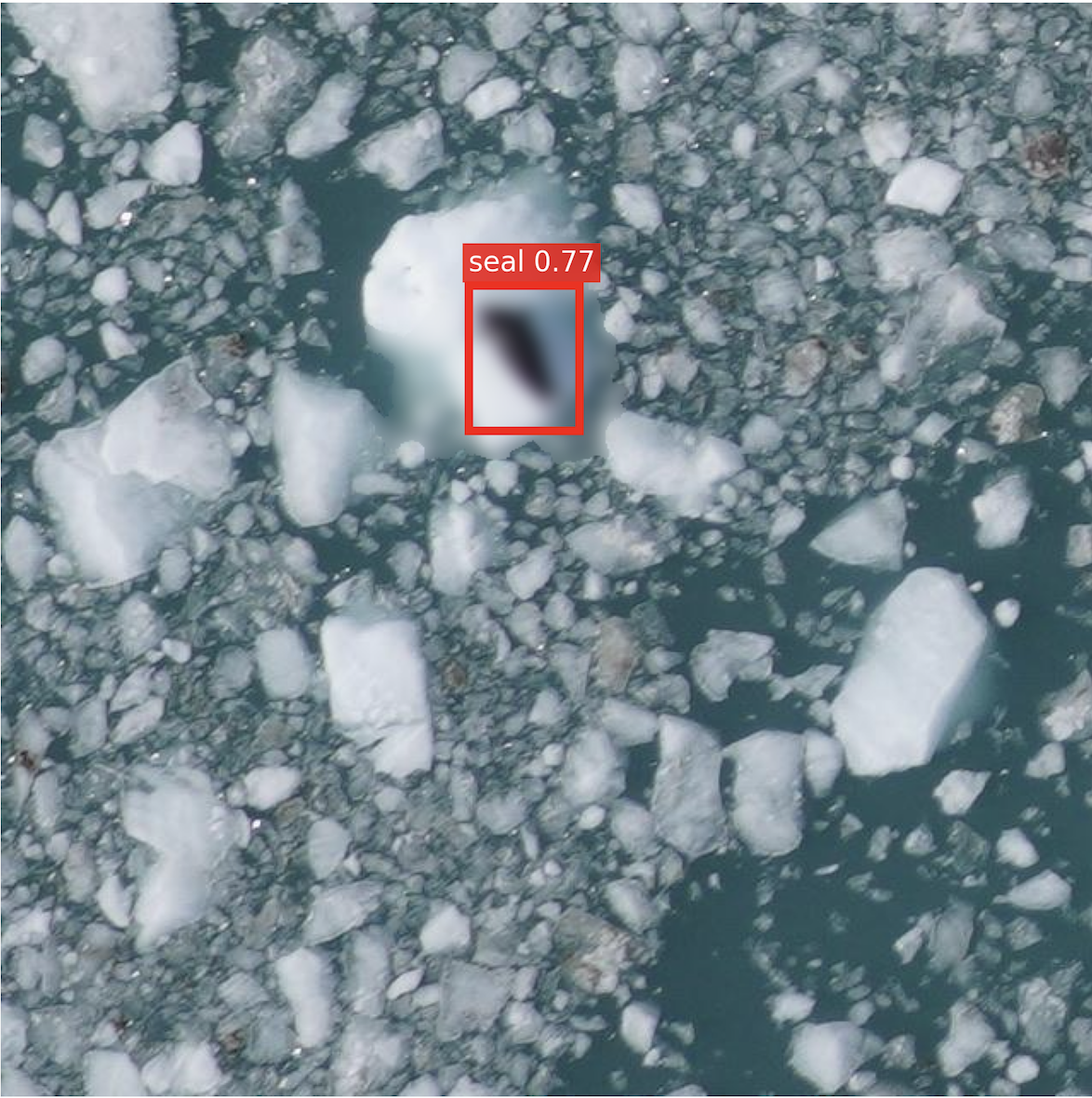}
        \caption{Blur}
    \end{subfigure}
  \caption{Perturbations of a successful seal detection. (a) Original detection with confidence score 1.00. (b) Masking the seal body leads to no detection. (c) Adding noise leads to no detection. (d) Blurring reduces confidence to 0.77 but preserves detection. These perturbations demonstrate how the model relies on seal contours and feature coherence for reliable predictions.}
  \vspace{-0.5em}
\end{figure}

Most false positives in the test set were attributable to annotation gaps, ambiguous background features, or visually confounding ice conditions. Identified errors included one missed seal annotation, four cases involving dark shapes at image edges, one dark shape in open water, one merged detection of two adjacent seals, and several instances of black ice. An example of a challenging failure (Figure 1, bottom row) illustrates a false positive in which the detector identifies a patch of black ice as a seal. In this sample, attribution methods provide important insight. Both LayerCAM and HiResCAM show strong activations over the black ice region, revealing that the model is relying on low-level visual similarity between seals and dark ice patches rather than on biological features. LIME confirms this by assigning this region high importance for the prediction. In these instances, attribution maps consistently highlighted the confounding structures, confirming the model’s attention was drawn to confounding structures rather than seals in these instances.

\section{Discussion}
 
This case study argues that post-hoc explainability should be standard practice when validating computer-vision models for ecological research and conservation monitoring. Our results highlight three axes important to the ecology domain: localization fidelity, faithfulness, and diagnostic utility. We show how explanations provide evidence for predictions while revealing practical limitations.

Our perturbation-based explanations provide a check of faithfulness: when putative evidence is removed or corrupted, we observe a reduction in our model’s confidence. Masking and noise produced large mean confidence drops with high flip rates, whereas blur yielded smaller changes. This indicates reliance on intact body contours and coherent feature patterns rather than contextual shortcuts, while also showing that coarse structural cues can sometimes sustain predictions. 

Explanations are valuable for diagnosing model limitations and guiding improvements (diagnostic utility). They indicate that the current detector may not generalize well in environments with large amounts of black ice or rock mixed with ice, as these features can be mistaken for seals. This analysis reveals key targets that humans should review when validating detections. Understanding this failure mode suggests potential solutions, such as targeted data augmentation to include more negative examples of black ice or refining the training dataset to better capture background variability. 

CAM variants, LIME, and perturbations offer complementary lenses on model behavior. Using multiple methods reduces over-reliance on any single explainer, reveals edge cases and artifacts for review, and strengthens confidence in the explanations when different methods consistently highlight the same visual features, providing a clearer basis for data curation and thresholding decisions.

As a case study, our aim is to demonstrate a practical workflow and its decision-supporting value, not to claim habitat- or model-general conclusions. All explainers are post-hoc approximations with method-specific sensitivities (i.e., CAM to layer choice/upsampling; LIME to superpixel segmentation and perturbation kernel), and our deletion edits (mask, noise, blur) can induce distribution shift that conflates causal evidence removal with artifacts. Faithfulness is measured via changes in post-NMS confidence, which may be miscalibrated and need not correspond to error rates or mAP changes. Localization is evaluated against bounding boxes rather than pixel masks, potentially penalizing explanations that focus on salient subregions or spill slightly outside labels. 

Together, these findings support our central claim: pairing object detection with post-hoc explainability moves beyond “black-box” outputs toward auditable evidence that improves trust and guides model improvement in conservation monitoring.

\bibliographystyle{plainnat}
\bibliography{neurips_2025}

\appendix
\section{Data Collection and Labeling Methods}
Drone surveys were conducted using a Wingtra One Gen II fixed-wing platform (Zurich, Switzerland) with vertical takeoff and landing (VTOL) capabilities imaging with a Sony Alpha 6100 APS-C camera (Tokyo, Japan) with a Sony E 20mm f/2.8 lens. Flight plans were created and carried out using WingtraPilot flight planning software. Flights operated at ~60–85 m altitude and ~9–22 m/s airspeed over regions that were historically sampled by occupied aircrafts. Drone operations were conducted under permit by NOAA and the NPS.

Flights in glacial fjords occurred along non-overlapping parallel transects oriented lengthwise through the glacial end of the fjord, with transects oriented approximately perpendicular to the glacier terminus. These flight plans were similar to those of historic surveys that used occupied aircrafts \citep{womble2020calibrating}, but were optimized to achieve high-density coverage of the inner regions of the fjords where seal densities are highest. Surveys from occupied aircrafts historically sampled the entire west arm of JHI along the same 12 transects year after year, maintaining a ~100-m buffer between photographs across transects and a ~20-m buffer between consecutive photographs along transects. Surveys from unoccupied aircrafts in 2023 and 2024 surveyed smaller gross extents with a series of nearly contiguous but not overlapping transects, maintaining a ~5-m buffer between photographs across transects and 65–70\% overlap between consecutive photographs along transects. Surveys from unoccupied aircrafts in glacial fjords consisted of 1–3 flights each using impromptu flight plans informed by the extent of floating ice habitat and drone performance in the prevailing weather conditions at the time of the survey. Surveys of terrestrial sites also consisted of parallel transects arranged in a high density to achieve a target of ~60\% overlap between photographs along transects and across transects.

 The training dataset was visually reviewed and manually thinned to remove photographs that did not include at least one positive instance of a harbor seal on floating ice. This was done to mitigate the risk of negative bias in model training, which can occur with an overabundance of negative training data. The resulting dataset images were each subdivided into tiles of 640×640 pixels. Each tile was manually inspected and annotated to mark all harbor seal locations using Labelme \citep{russell2008labelme}.

\section{Object Detection Model}

We employed a transfer learning approach for seal detection. Our model, a Faster R-CNN \citep{ren2016fasterrcnnrealtimeobject} with a ResNet-50 backbone and Feature Pyramid Network (FPN), was initialized with weights pre-trained on the COCO dataset. We then performed full fine-tuning, updating all layers of the network to adapt the model to our seal dataset. 
Prior to training, images were pre-processed to ensure compatibility with the Faster R-CNN model. The dataset was split into 1,744 training images, 151 validation images, and 76 test images of pinnipeds collected from aerial drone surveys. To improve robustness, the training dataset was additively augmented with geometric transformations (horizontal and vertical flips, rotations 0 - 45°, and random crops) and color adjustments (brightness, contrast, saturation, and hue). Validation images were processed only with normalization and tensor conversion for consistency, while the test set was completely held out and used only for final evaluation without any data augmentation. 
The model was trained using stochastic gradient descent with gradient clipping, warmup learning rate scheduling, and early stopping. Weights and Biases \citep{wandb} was used for experiment tracking.
Hyperparameters were tuned over learning rate (0.001-0.01), momentum (0.85–0.95), and weight decay (0.0001–0.001) using a grid search. The chosen configuration used a batch size of 8, a learning rate of 0.0090, momentum of 0.874, and weight decay of 0.0001. Although training was set for up to 200 epochs, the best model was obtained at epoch 67, where validation performance peaked with a mean Average Precision (mAP, averaged over IoU thresholds from 0.5 to 0.95) of 0.61, mAP50 of 0.95, and mean Average Recall (mAR, recall averaged across up to 100 detections per image) of 0.69. The corresponding training and validation losses were 0.11 and 0.13, respectively. On the held-out test set of 76 images, the final model achieved an overall mAP of 0.65, with mAP50 of 0.98 and mAP75 of 0.78. Performance was consistent across object (seal) sizes, with mAP values of 0.62 for small objects, 0.65 for medium, and 0.77 for large, while mAR100 reached 0.72. 

\section{Explainability Approaches}
\subsection{Gradient-based Class Activation Mapping}
Gradient-based CAM methods explain a model's prediction by backpropagating the gradient of a scalar target (e.g., the class logit \(y_c\)) to an internal convolutional layer and converting those gradients and activations into a spatial relevance map \citep{selvaraju2016gradcam}. The map is then upsampled to input resolution and visualized as a heatmap over the image. We apply two CAM variants: HiResCAM \citep{draelos2021usehirescaminsteadgradcam} and LayerCAM \citep{Jianglayercam}. Compared with traditional approaches Grad-CAM/Grad-CAM++ \citep{selvaraju2016gradcam}, both methods avoid global spatial pooling and instead use element-wise interactions between activations and gradients, yielding higher spatial fidelity—important for aerial imagery where animals are often small, low-contrast, or partially occluded.

To apply CAM, we first select a convolutional layer that preserves spatial structure, typically the last conv block of the detector's backbone/head. For Faster R-CNN, this is \texttt{backbone.body.layer4}. Let its feature tensor be \(A \in \mathbb{R}^{K \times U \times V}\), where \(A_k\) is the \(k\)-th channel and \((U,V)\) are spatial dimensions. For each prediction, we define the target scalar as the detector's class score/logit for the predicted class \(y_c\) and compute the gradient
\(G = \frac{\partial y_c}{\partial A} \in \mathbb{R}^{K \times U \times V}\).
These per-location derivatives indicate how sensitive \(y_c\) is to changes in \(A\).
We then form relevance using element-wise activation--gradient products, as shown in Eq.~\eqref{eq:layercam} (LayerCAM) and Eq.~\eqref{eq:hirescam} (HiResCAM).

\begin{align}
R_k &= \mathrm{ReLU}(G_k) \odot A_k, &
L_c &= \sum_{k=1}^{K} R_k
\label{eq:layercam}\\[4pt]
L_c &= \mathrm{ReLU}\!\left(\sum_{k=1}^{K} A_k \odot G_k\right)
\label{eq:hirescam}
\end{align}

We bilinearly upsample \(L_c\) to the input size, perform min--max normalization to \([0,1]\), and overlay the heatmap on the image for interpretation.
 
\subsection{Local Interpretable Model-Agnostic Explanations (LIME)}

Local Interpretable Model-agnostic Explanations (LIME) explains predictions by generating perturbed inputs and fitting a sparse local surrogate model to approximate the black-box decision boundary \citep{ribeiro2016whyitrustyoulime}. For images, the input is first segmented into superpixels, which are selectively masked and passed through the detector; the surrogate model then assigns importance weights to each segment, indicating its contribution to the prediction. Let the input image be segmented into regions \(S = \{s_1, \dots, s_n\}\). For each perturbed sample \(x' \subseteq S\), the model prediction is recorded as \(f(x')\). LIME solves a locally weighted regression problem to estimate coefficients \(\beta_i\) such that:
\[
f(x') \approx \beta_0 + \sum_i \beta_i \cdot \mathbf{1}[s_i \in x'],
\]
where \(\beta_i\) captures the relevance of segment \(s_i\).

We adapt LIME for multi-instance object detection with detector-aware weighting and proximity suppression. Our implementation introduces three refinements. First, we apply improved segmentation by converting images to the \textsc{LAB} color space and using SLIC \citep{scikitimage} with higher compactness and smoothing, producing more perceptually coherent regions. We then remove very small or black segments to avoid spurious explanations. Second, we adapt LIME to handle multiple detected objects. Given a set of instances \(I = \{i_1, \dots, i_M\}\), each with bounding box \(b_i\), confidence score \(c_i\), and area \(a_i\), we assign weights
\[
w_i =
\begin{cases}
\dfrac{c_i}{\sum_{j=1}^{M} c_j}, & \text{confidence mode},\\[6pt]
\dfrac{a_i}{\sum_{j=1}^{M} a_j}, & \text{area mode},\\[6pt]
\dfrac{1}{M}, & \text{uniform mode},
\end{cases}
\]
and compute combined instance scores using Intersection-over-Union \(\operatorname{IoU}\)–based matching. The confidence mode prioritizes objects with high prediction scores \(c_i\), focusing the analysis on features that drive the model’s most certain detections. The area mode weights objects by their bounding-box size \(a_i\), emphasizing the visual importance of larger instances. The uniform mode assigns equal weight \((1/M)\) to all detections, producing a balanced, aggregated explanation of feature importance for the class as a whole, unbiased by confidence or size.

We then construct an explanation map by merging segment importances with spatial weighting: segments overlapping or close to a detection box receive higher relevance, while distant or background regions are suppressed. This yields cleaner, less cluttered maps that focus attention on the true object instances, adapting LIME’s perturbation paradigm to the multi-instance detection setting.

\subsection{Perturbation-based Explanations}
For perturbation-based (deletion) explanations, we begin by processing an image \(x\) with a detector \(f\), which produces detections \(D_x=\{(b_k,c_k,s_k)\}_{k=1}^{K}\), where each detection has a bounding box \(b_k\), class \(c_k\), and confidence \(s_k\). For a chosen target instance \(t=(b_t,c_t,s_t)\), the goal is to construct an edited image \(x'\) such that the target either disappears or its confidence falls below a fixed threshold \(\tau=0.5\), while edits remain minimal.

First, we localize the region to be perturbed. The image is segmented into superpixels, and we take those intersecting the target’s bounding box. If this set is too small, we expand it by including a thin dilated ring around the box (\(2\%\) of the image, at least \(2\) pixels wide).

Next, we apply perturbations within these regions. Each perturbation is applied on a mask dilated by \(2\) pixels to ensure coverage. The options are: constant background fill, Gaussian blur \((\sigma=5)\), and additive noise (level \(\approx 0.6\)). Let \(x \odot P_{S}\) denote applying perturbation \(P\) to \(x\) on the superpixel set \(S\).

In the additive noise condition, pixels within the perturbed region are replaced with random values sampled from a uniform distribution \(U(0,1)\) and rescaled to the original image range. This operation preserves global brightness statistics while destroying local semantic structure, serving as a standard information-removal baseline in faithfulness testing \citep{ivanovs2021perturbation}.

Then, at each iteration, we use a greedy forward-selection procedure: we perturb the superpixel that yields the largest reduction in the target’s confidence. Formally, the set of perturbed superpixels is updated as
\[
S_{t+1}
= S_t \cup
\operatorname*{arg\,min}_{s \in R \setminus S_t}\;
f\!\left(x \odot P_{\,S_t \cup \{s\}}\right),
\]
where \(R\) is the eligible region, \(P\) the perturbation operator, and \(f\) the detector’s confidence score.

The process continues until either the target’s confidence drops below
\[
f\!\left(x \odot P_{\,S_t}\right) < \tau,
\]
or the maximum number of iterations (default \(80\)) is reached.

Finally, pre- and post-edit detections are matched by class agreement and Intersection-over-Union (IoU). If no detection overlaps the original target by more than \(\delta=0.2\), the target is considered absent and its confidence is set to zero. A perturbation is valid if the matched confidence falls below \(\tau\). Minimality is quantified by the perturbed area fraction.

We use the three commonly applied perturbations in explainability research to assess faithfulness (masking, additive uniform noise, and Gaussian blur). The three are representative of the most common ways to remove information that have been used in both deletion–insertion and occlusion based testing in computer vision \citep{ivanovs2021perturbation}. Our goal is not to propose new perturbations but to compare these established methods in the context of ecologically valid object detection, quantifying how each affects model confidence when evidence for object presence is removed.

\section{Assessing Explanations Along Ecologically-Relevant Axes}

\subsection{Localization fidelity}
We quantified how well explanation methods highlight the animal rather than the surrounding context using LayerCAM and HiResCAM attribution maps. We calculated the proportion of mid- to high-attribution pixels falling within annotated bounding boxes, normalized by the total number of attribution pixels, which we call the attribution ratio. In addition, we measured the percentage of images in which the maximum-saliency pixel was located inside a labeled box, which we call the maximum saliency hit-rate. 

\subsection{Faithfulness}
To evaluate faithfulness, we utilize our perturbation-based explanations and compute two criteria. The flip rate measures how often the target is suppressed below a threshold \(\tau\),
\[
\mathrm{FR}_p \;=\; \frac{1}{N}\sum_{i=1}^{N} \mathbf{1}\!\left[\, s_i^{\prime (p)} < \tau \,\right],
\]
where \(s_i\) is the original confidence and \(s_i^{\prime (p)}\) the confidence after perturbation \(p\).

The confidence drop is defined as
\[
\triangle s_i^{p} \;=\; s_i - s_i^{\prime (p)},
\]
and we report both the mean across all cases,
\[
\mathrm{CD}_p \;=\; \frac{1}{N}\sum_{i=1}^{N} s_i^{p},
\]
and the conditional mean among unsuccessful flips \(\big(s_i^{\prime (p)} \ge \tau\big)\):
\[
U_p \;=\; \{\, i : s_i^{\prime (p)} \ge \tau \,\}, \qquad
\mathrm{CD}_{\text{flip}}^{p} \;=\; \frac{1}{|U_p|}\sum_{i \in U_p} s_i^{p}.
\]
Together, these metrics capture both the efficacy of perturbations in flipping predictions and the extent to which confidences are reduced even when flips do not occur.

\subsection{Diagnostic Utility}
Explanations were reviewed for their capacity to identify conditions in which the model is prone to error, thus forming opportunities to improve both model design and dataset quality. We conducted a targeted review of false positives (detections not annotated as seals) by overlaying predicted bounding boxes and saliency maps with ground-truth labels. False positive detections were isolated and manually inspected to characterize the contexts in which they occurred. This analysis was performed across LayerCAM, HiResCAM, and LIME to confirm the consistency of observed patterns.

\section{Black Masking Perturbation}

We also conducted black masking experiments (RGB: 0,0,0) on both YOLO and Faster R-CNN models using the same 76 test images, following the same perturbation approach.

\subsection{Results}

For the YOLO Model, 44 out of 76 images (57.9\%) produced successful counterfactuals with a 100\% flip rate, requiring an average of 1.2 $\pm$ 0.4 superpixels and changing 0.45\% $\pm$ 0.20\% of the image area. The mean confidence drop was 0.73 $\pm$ 0.15. For the Faster R-CNN Model, 40 out of 76 images (52.6\%) produced successful counterfactuals with a 100\% flip rate, requiring an average of 1.3 $\pm$ 0.5 superpixels and changing 0.52\% $\pm$ 0.25\% of the image area. The mean confidence drop was 0.75 $\pm$ 0.18.

\subsection{Comparison with Other Perturbation Methods}

Black masking achieved a 100\% flip rate, exceeding the effectiveness of mean background masking (97.33\%), noise (80.00\%), and blur (21.33\%).

These results indicate that the model does not rely primarily on color cues for detection: even when regions were replaced with uniform black pixels, predictions were suppressed as effectively as when using the mean or noisy masks. This indicates that the model’s decisions depend more on structural and spatial features than on specific color information.

\end{document}